\documentclass[11pt]{article}

\usepackage[final]{acl}

\usepackage{times}
\usepackage{latexsym}

\usepackage[T1]{fontenc}

\usepackage[utf8]{inputenc}

\usepackage{microtype}

\usepackage{inconsolata}

\usepackage{graphicx}

\usepackage{amsmath}
\usepackage{amssymb}
\usepackage{booktabs}
\usepackage{multirow}
\usepackage{enumitem}
\usepackage{comment}
\usepackage{xcolor}
\usepackage{placeins}
\usepackage{tikz}
\usetikzlibrary{positioning, arrows.meta, calc, fit, shapes.arrows}
\usepackage{pifont}
\usepackage[breakable]{tcolorbox}
\usepackage{xspace}

\definecolor{promptbrown}{HTML}{8b4513}
\tcbset{
  promptbox/.style={
    colback=gray!5!white,
    colframe=promptbrown!75!black,
    boxrule=0.3mm,
    width=\textwidth,
    arc=3mm,
    auto outer arc=true,
    fonttitle=\bfseries\small,
    fontupper=\small
  }
}

\newcommand{\StatFormBench}{\textsc{StatFormBench}\xspace}
\newcommand{\smalldisplay}{%
  \edef\savedbaselineskip{\the\baselineskip}%
  \small
  \baselineskip=\savedbaselineskip\relax
}

\title{Benchmarking Language Models for Statistical Problem Formulation}

\author{
    Chen Wang\textsuperscript{1}\footnotemark[1], 
    Junzhe Zhao\textsuperscript{1}\footnotemark[1], 
    Xin Cong\textsuperscript{1}\footnotemark[2], 
    Wanlu Deng\textsuperscript{1}, 
    Ke Deng\textsuperscript{1} \\  
    \textbf{\textsuperscript{1}} Department of Statistics and Data Science, Tsinghua University \\
    \texttt{\small \{wangchen\_23, zhaojz26\}@mails.tsinghua.edu.cn} \\
    \texttt{\small \{congxin1995, wanludeng, kdeng\}@tsinghua.edu.cn}
}

\begin{document}
\maketitle

\renewcommand{\thefootnote}{\fnsymbol{footnote}}
\footnotetext[1]{\ \ Equal contribution.}
\footnotetext[2]{\ \ Corresponding author.}
\renewcommand{\thefootnote}{\arabic{footnote}}

\begin{abstract}

Large language models~(LLMs) are increasingly used as assistants for statistical and data science work, yet existing evaluations largely assume the analysis target is already specified. In practice, users arrive with informal goals and heterogeneous data, leaving the model to decide what statistical task is implied and which data are relevant. We first formalize this upstream step as Statistical Problem Formulation and decompose it into two subtasks: (1) Statistical Problem Classification and (2) Variable Identification \& Role Assignment. We then introduce \StatFormBench, a benchmark built from five cross-domain statistics textbooks and a data science case library, covering diverse problem types, data representations, and scenario styles. It contains 1,013 samples spanning 20 coarse-grained and 85 fine-grained statistical problem categories. Across 14 open- and closed-source LLMs, the best zero-shot models reach only 72.0 fine-grained classification accuracy and 63.2 variable set overlap. No model performs consistently best across the two subtasks, while enhanced prompting strategies yield only limited or inconsistent gains.
We release the benchmark data on Hugging Face at \url{https://huggingface.co/datasets/THU-CongLab/StatFormBench} and the evaluation code on GitHub at \url{https://github.com/THU-CongLab/StatFormBench}.

\end{abstract}

\section{Introduction}
\label{sec:introduction}
\begin{figure*}[t]
\centering
\includegraphics[width=\textwidth]{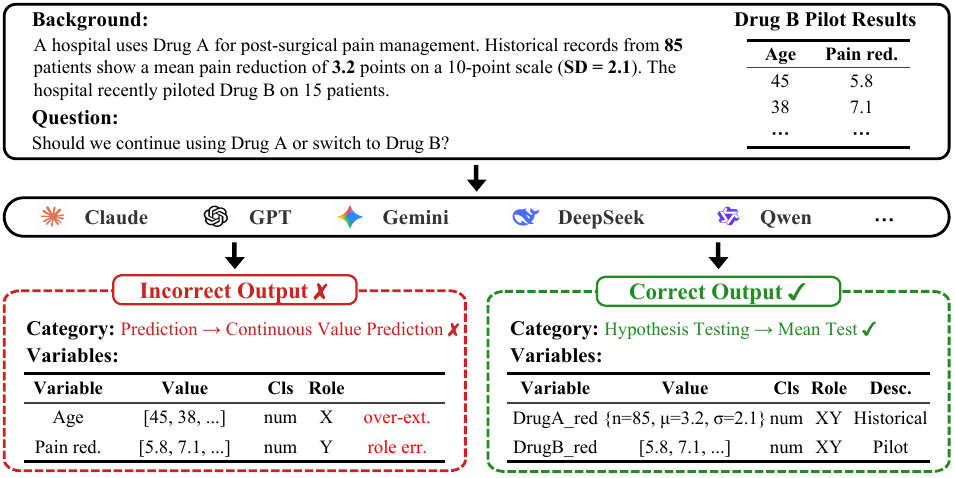}
\caption{Overview of statistical problem formulation.
Given a scenario with heterogeneous data, the model must classify the statistical problem type and extract the relevant variables with their roles.
}
\label{fig:task_overview}
\end{figure*}

Large language models~(LLMs) are increasingly expected to serve as general-purpose assistants for data science and statistical analysis work, helping users inspect datasets, choose analyses, and answer questions grounded in data~\citep{dacode,datascibench,scienceagentbench}. Much of this work assumes that the analysis target has already been made explicit, for example by specifying a prediction task, a hypothesis test, or a modeling objective~\citep{stateval,qrdata,dacode,datascibench}. 
In practice, however, users often begin with informal goals rather than formal analysis specifications. As illustrated in Figure~\ref{fig:task_overview}, even a simple practical request may require the model to distinguish the intended hypothesis test from a superficially plausible prediction task. It also needs to determine which data objects are required for that test, rather than treating all available fields as relevant. These implicit choices determine whether the subsequent analysis is statistically valid. We refer to this upstream specification step as statistical problem formulation.

We define \textbf{Statistical Problem Formulation} as the task of mapping an informal, natural-language analysis request to (i) the type of statistical problem it implies and (ii) the relevant data objects together with their analytical roles. These two components correspond to two closely connected decisions. The first, \textbf{Statistical Problem Classification}, determines what type of statistical task a scenario calls for, such as comparing group means, examining associations between variables, or predicting future outcomes. The second, \textbf{Variable Identification and Role Assignment}, determines which data objects are relevant to the task and what analytical role each object plays, such as an independent variable or a dependent variable. Once these decisions have been made, the subsequent choice of statistical methods becomes more constrained. By contrast, formulation itself often depends on interpreting the user's intent, the available data, and the surrounding domain context. Whether LLMs can reliably perform this step remains largely unexplored.

Recent benchmarks have examined LLMs in statistics and data science, covering statistical knowledge and reasoning~\citep{stateval,qrdata,climateviz} as well as data science execution~\citep{dacode,datascibench,scienceagentbench}. These studies provide useful evidence about whether LLMs can reason with statistical concepts or carry out specified analyses, but they generally assume that the analysis target is already given or strongly implied. As a result, they do not directly test whether a model can first formulate the statistical task from an informal real-world request.

Evaluating this capability is challenging because formulation differs from standard question answering or task execution in several ways. First, the input is often not written in statistical language: users may describe practical goals with ambiguity, domain-specific wording, and distracting contextual details. Second, the relevant data may be represented in heterogeneous forms, including raw tabular variables, summary statistics such as means and standard deviations, contingency tables, and other structured or semi-structured objects. Third, the target label space is broad and weakly delimited. Deciding whether a scenario calls for comparison, association analysis, prediction, reliability assessment, or another statistical task requires jointly interpreting the user's objective, the data structure, and the domain context. The same problem-level understanding is also needed to separate relevant variables from incidental information and to assign their analytical roles.

To this end, we introduce \StatFormBench, a benchmark for statistical problem formulation in realistic, unstructured settings. \StatFormBench operationalizes formulation through two tasks aligned with the decisions above. \textbf{Statistical Problem Classification} asks a model to identify the statistical problem type expressed by a scenario, and \textbf{Variable Identification \& Role Assignment} asks it to identify the data objects relevant to the task and assign their analytical roles. The benchmark is constructed from statistics textbook exercises across applied domains and a data science case library reflecting diverse business scenarios, allowing us to cover both controlled pedagogical problems and more open-ended real-world analysis requests.
Each sample is built through a unified pipeline that extracts raw triples of question, answer, and data, splits multi-part problems into individual instances, filters for quality, and annotates problem types and variable roles. The problem descriptions are then rewritten into realistic scenario-style language.
We evaluate 14 open-source and proprietary LLMs. The best zero-shot fine-grained classification accuracy is 72.0, while the highest variable set overlap is 63.2, with the two results achieved by different models. Fine-grained analysis further shows that performance varies substantially across problem categories and data sources, while standard prompting strategies yield limited and inconsistent gains.

Our contributions are as follows:
(1) we are the first to study statistical problem formulation as an evaluation problem for LLMs, decomposing it into statistical problem classification and variable identification \& role assignment;
(2) we construct the \StatFormBench benchmark covering diverse statistical problem types, data representations, and scenario styles, with samples derived from both textbooks and real-world data science cases;
(3) we evaluate 14 open-source and proprietary LLMs, showing that no model performs consistently best across formulation tasks and that current models still struggle to map informal requests to precise statistical formulations regardless of the prompting strategy used.

\section{Related Work}
\label{sec:related_work}

\subsection{LLMs for Statistics and Data Science}
\label{sec:rw_stat}
Recent benchmarks evaluate LLMs for statistics and data science mainly along two lines. Both are closely related to our setting, but these studies generally assume that the analytical goal is already specified or implied.

\paragraph{Statistical knowledge and reasoning.}
The first line studies whether LLMs can reason with statistical concepts once the analytical goal is explicit or strongly implied. StatQA~\citep{statqa} evaluates method applicability and variable selection for pre-formulated questions over structured tables, StatEval~\citep{stateval} extends this direction to foundational exercises and research-level proof tasks, and QRData~\citep{qrdata} and ClimateViz~\citep{climateviz} examine statistical or causal reasoning over data sheets and scientific charts.
These benchmarks assess statistical competence in settings where the target analytical goal is already defined, whereas we focus on inferring the statistical problem itself and identifying the data objects required for solving it.

\paragraph{Data science execution.}
The second line evaluates LLMs as executors of specified analytical tasks. DA-Code~\citep{dacode} and DataSciBench~\citep{datascibench} focus on code generation for data preprocessing, analysis, and modeling, while ScienceAgentBench~\citep{scienceagentbench}, LMR-Bench~\citep{lmrbench}, DSBench~\citep{dsbench}, and BLADE~\citep{blade} extend evaluation to data-driven science, ML research reproduction, and realistic data science workflows. These works increasingly consider intermediate decisions, but they still largely assume that research questions or analytical objectives have been given.

\subsection{LLMs for Problem Formulation}
\label{sec:rw_formulation}

\paragraph{Optimization modeling.}
Another line of work evaluates whether LLMs can map natural-language descriptions to formal problem representations. Optimization modeling is the most developed example. NL4Opt~\citep{nl4opt} and subsequent benchmarks such as OptiMUS~\citep{optimus}, Chain-of-Experts~\citep{chainofexperts}, ORLM~\citep{orlm}, LLMOPT~\citep{llmopt}, and OptiBench~\citep{optibench} treat formulation as an explicit target rather than a hidden precursor to solving.

\paragraph{Mathematical and domain-specific formulation.}
Related benchmarks extend this formulation perspective to broader domains. ModelingAgent~\citep{modelingagent} and MM-Agent~\citep{mmagent} study mathematical modeling from competition problems, MedCalc-Bench~\citep{medcalcbench} and MedRaC~\citep{medrac} evaluate clinical calculation through formula selection, entity extraction, and computation, and DiscoveryBench~\citep{discoverybench} evaluates data-driven hypothesis generation in scientific discovery through context, variables, and relationships.
Together, these studies show the value of evaluating intermediate abstractions rather than only final answers.

Statistical problem formulation shares this mapping challenge, but is less anchored by a fixed schema. Unlike optimization formulation, which largely amounts to instantiating a fixed template of elements such as objectives and constraints, statistical formulation must infer the analytical question from domain-specific descriptions and select from a large, heterogeneous, and weakly delimited space of problem types. It must also identify diverse data objects, ranging from raw variables to grouped summaries and contingency tables. To our knowledge, no existing benchmark systematically evaluates this upstream formulation ability.

\begin{figure*}[!t]
\centering
\includegraphics[width=\textwidth]{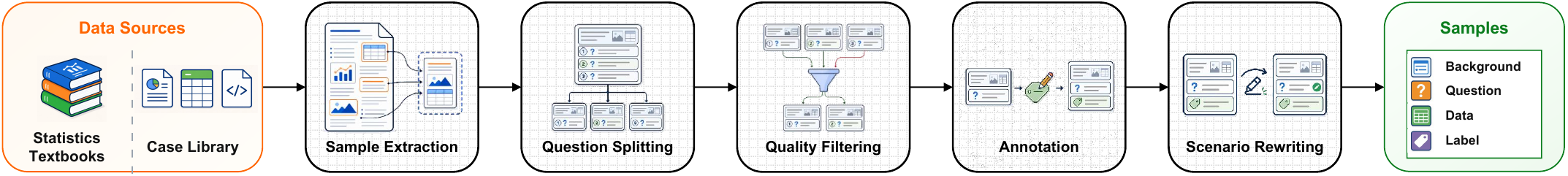}
\caption{Overview of our \StatFormBench benchmark construction pipeline.}
\label{fig:pipeline}
\end{figure*}

\section{Task Formulation}
\label{sec:task_formulation}
We decompose statistical problem formulation into two tasks defined over an input triple $(B,Q,D)$, where $B$ describes the practical background, $Q$ states the user request, and $D$ refers to data. $D$ varies in both form and content (e.g., tables, textual descriptions, separate files, etc.). In content, it ranges from raw observations to summary statistics and contingency tables.

\subsection{Task 1: Statistical Problem Classification}
Given $(B,Q,D)$, the task is to predict
{\smalldisplay
\begin{equation}
c = (c_{\mathrm{CG}}, c_{\mathrm{FG}}),\quad c_{\mathrm{CG}}\in\mathcal{C}_{\mathrm{CG}},\; c_{\mathrm{FG}}\in\mathcal{C}_{\mathrm{FG}},
\end{equation}
}
where $c$ denotes the statistical problem type. The label space is a two-level hierarchy. $\mathcal{C}_{\mathrm{CG}}$ captures coarse analytical intent, such as descriptive statistics, classification and prediction, and hypothesis testing and interval estimation, while $\mathcal{C}_{\mathrm{FG}}$ subdivides each first-level category into specific problem types. We classify the problem type rather than the solving method because the same statistical problem can admit multiple methods, and committing to one method would blur the distinction between problem identification and method selection.

\subsection{Task 2: Variable Identification and Role Assignment}
Given the same input, the task is to produce a set $V=\{v_1,\ldots,v_k\}$ of variables required for solving $Q$, where each variable object is represented as
{\smalldisplay
\begin{equation}
v_i = (\mathrm{id}_i,\;\mathrm{desc}_i,\;\mathrm{value}_i,\;\mathrm{role}_i).
\end{equation}
}
Here $\mathrm{id}_i$ is a short identifier, typically reusing a name from $D$. $\mathrm{desc}_i$ describes its domain meaning, and $\mathrm{value}_i$ stores the corresponding values from $D$. The role field specifies how $v_i$ participates in solving $Q$ and takes one of four values: \texttt{predictor}, \texttt{response}, \texttt{both}, or \texttt{N/A}. The roles \texttt{predictor} and \texttt{response} are assigned when the problem implies a directional relationship. The role \texttt{both} is used when variables are related without a required direction, as in correlation analysis, or when a variable plays both directional roles. The role \texttt{N/A} marks variables that are needed but have no directional role, such as identifiers, filter variables, and inputs to unsupervised analyses. Notably, roles are question-conditional, since the same data may play different roles under different user requests.

\begin{figure*}[!t]
\centering

\includegraphics[width=0.9\textwidth]{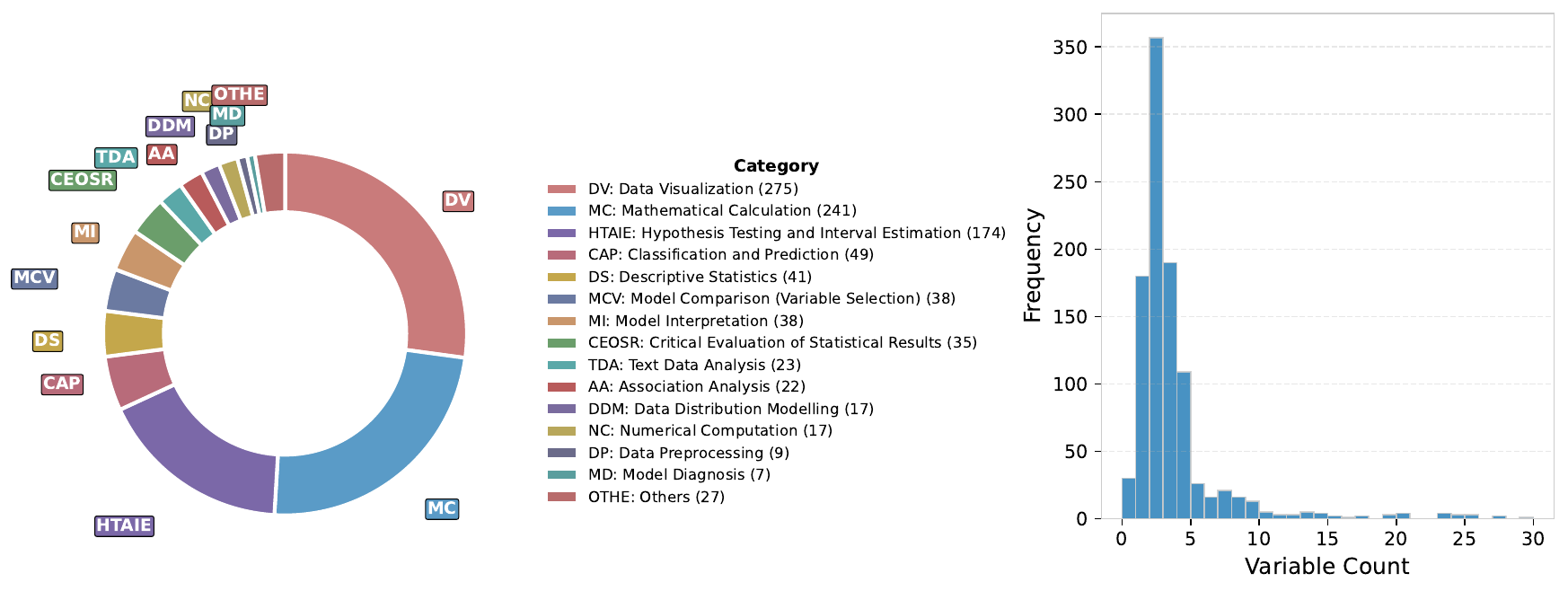}
\caption{Dataset statistics. \textbf{Left}: Distribution of samples across coarse-grained categories. \textbf{Right}: Distribution of the number of variables per sample.}
\label{fig:dataset_dist}
\end{figure*}

\section{Benchmark Construction}
\label{sec:benchmark}

\subsection{Data Sources}
\label{sec:data_sources}
We build \StatFormBench{} from two complementary sources. The first source consists of worked examples and exercises from five statistics textbooks spanning diverse applied domains, including public health~\citep{gerstman_biostatistics}, business and economics~\citep{mcclave_business_stats}, reliability analysis~\citep{pham_reliability}, environmental science~\citep{cook_environment_stats}, and clinical research~\citep{altman_medical_stats}. These samples provide broad coverage of canonical statistical problem types and usually include compact data in text or inline tables. The second source is a case-based data science library from GouXiongHui~\citep{gouxionghui}, where each case centers on a substantive analytical question, accompanied by an analysis report, structured datasets, and reference code. These samples provide more open-ended analytical contexts with larger data files in diverse formats.

\subsection{Construction Pipeline}
\label{sec:pipeline}

To enable uniform evaluation, both sources are converted into samples of the form $(B,Q,D,c,V)$, where $(B,Q,D)$ is the evaluation input, $c$ is the problem-type label, and $V$ is the reference variable set. Figure~\ref{fig:pipeline} summarizes the construction pipeline. We describe the main stages below with further details in Appendix~\ref{sec:appendix_data}.

\paragraph{Sample extraction and splitting.}
Raw materials are first parsed into entries $(Q^o,A^o,D)$, where the superscript $o$ denotes the original form. Specifically, $Q^o$ is the original problem statement, $A^o$ is the original reference answer, and $D$ contains the associated data. Entries with multiple sub-questions are then split into individual samples $(\tilde{B},\tilde{Q},\tilde{A},D)$ by extracting the shared background $\tilde{B}$ from $Q^o$ and pairing each sub-question $\tilde{Q}$ with its answer $\tilde{A}$.

\paragraph{Quality filtering.}
We remove samples that are too narrow to test formulation, such as value lookup and pure concept recall, and discard cases with incomplete information, such as questions that depend on the answer to a previous problem.

\paragraph{Initial label construction.}
Each retained sample is annotated with a category $c=(c_{\mathrm{CG}},c_{\mathrm{FG}})$ and a variable set $V=\{v_1,\ldots,v_k\}$. Domain experts trained in statistics at the undergraduate and doctoral level first construct an initial taxonomy from standard statistical practice, and the taxonomy is refined during annotation when new problem types arise. Specifically, the annotator LLM proposes candidate categories, and a domain expert decides whether to adopt or merge them. The final taxonomy contains $|\mathcal{C}_{\mathrm{CG}}|=20$ first-level and $|\mathcal{C}_{\mathrm{FG}}|=85$ second-level categories (Appendix~\ref{sec:appendix_taxonomy}). Variable labels are grounded in the solving process of $\tilde{A}$, so variables mentioned in the scenario but not needed for the solution are excluded. For the textbook subset, Claude Opus 4.6~\cite{claudeopus46} produces the initial annotations of both $c$ and $V$. For the case-library subset, separate GPT-4o~\cite{gpt4o} calls produce the initial annotations of $c$ and $V$. The model-generated labels serve only as initial candidates, and every final label is subsequently verified by human reviewers as described in Section~\ref{sec:human_verification}.

\paragraph{Scenario rewriting.}
As both sources are pedagogical in origin, the original $(\tilde{B},\tilde{Q})$ often use instructional or explicitly statistical language and rarely include information beyond what the problem requires. We therefore use GPT-5 to rewrite each problem from the perspective of a domain practitioner without statistical training. The rewrite preserves the analytical intent, numerical values, and named entities, while removing direct cues about the target statistical problem category, introducing minor contextual details not directly relevant to the solution, and rephrasing the request in more natural scenario-style language. The rewritten results $(B,Q)$, derived from $(\tilde{B},\tilde{Q})$, combine with $D$, $c$, and $V$ from earlier stages to form the final samples.

\paragraph{Source-specific adaptations.}
The textbook subset follows the pipeline above directly, with PDF extraction assisted by MinerU\footnote{\url{https://mineru.net/}} and manual proofreading. The case-library subset differs only in extraction. We retain cases with both a complete analysis report and a structured dataset, induce $(\tilde{B},\tilde{Q},\tilde{A})$ from the report through a multi-call LLM chain, and summarize raw data files into a uniform data description used as $D$. To avoid directly releasing the source data, we further replace the row-level values in $D$ with simulated values generated by GPT-5. Further details are provided in Appendix~\ref{sec:appendix_case_library}.

\subsection{Human Verification}
\label{sec:human_verification}
To ensure the reliability of the benchmark, all 1,013 samples in \StatFormBench{} are manually reviewed. The review covers the problem category $c$ and the variable set $V$, with both verified against the reference answers in the source materials.

For the 629 textbook samples, we conduct two rounds of review by different PhD-level experts in statistics. The first reviewer checks and corrects each initial annotation against the exercise and its reference answer. A second reviewer then re-examines the corrected annotation. Any disagreements between the two rounds are resolved by a senior reviewer, who determines the final labels. For the 384 case-library samples from GouXiongHui, the initial annotations are easier to verify because each sample is derived from a complete analysis report that documents the underlying analytical process. A PhD-level expert reviews every annotation against the report and associated structured data.

\subsection{Dataset Statistics}
\label{sec:dataset_stats}
The final benchmark contains 1,013 samples, with 629 from textbooks and 384 from the case library. As shown in Figure~\ref{fig:dataset_dist}, the dataset covers a broad range of problem categories, with substantial mass in data visualization, mathematical calculation, and hypothesis testing and interval estimation. The number of variables per sample is concentrated at small values but has a long tail, reflecting both compact textbook exercises and more complex case-library scenarios. A more detailed breakdown of the category distribution is provided in Appendix~\ref{sec:appendix_dataset_statistics}.

\section{Evaluation Metrics}
\label{sec:metrics}

For the $j$-th sample, let $c^{(j)}=(c_{\mathrm{CG}}^{(j)},c_{\mathrm{FG}}^{(j)})$ and $\hat{c}^{(j)}=(\hat{c}_{\mathrm{CG}}^{(j)},\hat{c}_{\mathrm{FG}}^{(j)})$ denote the reference and predicted problem categories, and let $V^{(j)}$ and $\hat{V}^{(j)}$ denote the reference and predicted variable sets.

\subsection{Metrics for Statistical Problem Classification}
We report coarse-grained and fine-grained classification accuracy,
{\smalldisplay
\begin{align}
\mathrm{ACC}_{\mathrm{CG}}&=\frac{1}{N}\sum_{j=1}^{N}\mathbf{1}[\hat{c}_{\mathrm{CG}}^{(j)}=c_{\mathrm{CG}}^{(j)}],\\
\mathrm{ACC}_{\mathrm{FG}}&=\frac{1}{N}\sum_{j=1}^{N}\mathbf{1}[\hat{c}_{\mathrm{FG}}^{(j)}=c_{\mathrm{FG}}^{(j)}].
\end{align}
}
Since a correct fine-grained prediction implies a correct coarse-grained one, the gap between the two reflects how often models identify the broad analytical intent but miss the specific problem type.

\subsection{Metrics for Variable Identification and Role Assignment}

\paragraph{Variable matching.}
Because variable identifiers are model-generated, we first match each predicted variable $\hat{v}\in\hat{V}^{(j)}$ to a reference variable $v\in V^{(j)}$ when their \texttt{value} fields are exactly equal. Let $M^{(j)}$ be the resulting set of matched pairs. Unmatched predictions and unmatched reference variables are treated as false positives and false negatives. Further implementation details and results from a human validation study of this matching rule are provided in Appendix~\ref{sec:appendix_exact_matching}.

\paragraph{Variable identification metrics.}
To evaluate variable identification, we report three set-level metrics. The Jaccard coefficient of variable sets (JCV) measures overall overlap,
{\smalldisplay
\begin{equation}
\mathrm{JCV}=\frac{1}{N}\sum_{j=1}^{N}\frac{|M^{(j)}|}{|\hat{V}^{(j)}|+|V^{(j)}|-|M^{(j)}|}.
\end{equation}
}
To further distinguish over-prediction from missed variables, we also report variable precision (PV) and variable recall (RV),
{\smalldisplay
\begin{align}
\mathrm{PV}&=\frac{1}{N}\sum_{j=1}^{N}\frac{|M^{(j)}|}{|\hat{V}^{(j)}|},\\
\mathrm{RV}&=\frac{1}{N}\sum_{j=1}^{N}\frac{|M^{(j)}|}{|V^{(j)}|}.
\end{align}
}

\paragraph{Variable role identification (VRI).}
Let $\mathrm{role}(v)$ denote the role field of $v$. We measure the fraction of reference variables that are both recovered and assigned the correct role using VRI,
{\smalldisplay
\begin{equation}
\mathrm{VRI}=\frac{1}{N}\sum_{j=1}^{N}\frac{\sum_{(\hat{v},v)\in M^{(j)}}\mathbf{1}[\mathrm{role}(\hat{v})=\mathrm{role}(v)]}{|V^{(j)}|}.
\end{equation}
}
This metric jointly penalizes missed variables and incorrect role labels.

\begin{table}[t]
\centering
\small
\resizebox{\linewidth}{!}{%
\begin{tabular}{lcccccc}
\toprule
\multirow{2}*{\textbf{Model}} & \multicolumn{2}{c}{\textbf{Classification}} & \multicolumn{4}{c}{\textbf{Variable Id.\ \& Role}} \\
\cmidrule(lr){2-3} \cmidrule(lr){4-7}
 & $\mathrm{ACC}_{\mathrm{CG}}$ & $\mathrm{ACC}_{\mathrm{FG}}$ & JCV & PV & RV & VRI \\
\midrule
Claude Opus 4.6          & \underline{77.7} & \underline{69.5} & \textbf{63.2} & \underline{65.8} & \textbf{72.9} & \textbf{66.2} \\
Claude Sonnet 4.6        & 71.9 & 63.9 & \underline{62.4} & 65.0 & \underline{70.9} & 63.3 \\
GPT-5.5                  & 72.9 & 67.2 & 52.1 & 53.7 & 67.2 & 61.1 \\
GPT-5.4                  & 75.2 & 65.8 & 52.0 & 53.7 & 68.3 & 62.0 \\
Gemini 3.1 Pro           & \textbf{79.0} & \textbf{72.0} & 61.5 & 64.5 & 68.3 & 63.1 \\
Gemini 3 Flash           & 77.0 & 68.4 & \underline{62.4} & \textbf{66.1} & 68.9 & 63.6 \\
\midrule
Kimi-K2.5                & 71.5 & 64.8 & 59.3 & 61.3 & 69.8 & 63.8 \\
DeepSeek-V4-Pro          & 66.2 & 59.1 & 61.5 & 64.9 & 67.3 & 60.3 \\
DeepSeek-V4-Flash        & 71.2 & 64.8 & 54.5 & 56.1 & \underline{70.9} & \underline{65.3} \\
Qwen3.5-397B-A17B        & 70.6 & 64.6 & 57.2 & 60.5 & 66.5 & 61.4 \\
Qwen3.5-9B               & 29.1 & 20.3 & 48.3 & 51.6 & 58.1 & 45.0 \\
Qwen3.5-4B               & 37.7 & 30.6 & 45.4 & 48.7 & 54.9 & 41.1 \\
Qwen3.5-2B               & 7.7 & 3.5 & 22.3 & 32.0 & 25.2 & 7.1 \\
Qwen3.5-0.8B             & 7.9 & 3.8 & 5.6 & 8.3 & 7.6 & 3.0 \\
\bottomrule
\end{tabular}
}
\caption{Main results on \StatFormBench{}. \textbf{Bold} denotes the best result, and \underline{underline} denotes the second best.}
\label{tab:main_results}
\end{table}

\section{Experiments}
\label{sec:experiments}

\subsection{Setup}
\label{sec:setup}

We evaluate 14 LLMs from six model families on the two tasks defined in Section~\ref{sec:task_formulation}. The closed-source group includes Claude Opus 4.6~\cite{claudeopus46}, Claude Sonnet 4.6~\cite{claudesonnet46}, GPT-5.5~\cite{gpt5.5}, GPT-5.4~\cite{gpt5.4}, Gemini 3.1 Pro~\cite{gemini31pro}, and Gemini 3 Flash~\cite{gemini3flash}. The open-source group includes Kimi-K2.5~\cite{kimiteam2026kimik25visualagentic}, DeepSeek-V4-Pro, DeepSeek-V4-Flash~\cite{deepseekai2026deepseekv4}, and five Qwen3.5 variants (397B-A17B, 9B, 4B, 2B, and 0.8B)~\cite{qwen3.5} that cover a broad range of model scales. All models are evaluated on the full benchmark under zero-shot prompting, and additional prompting strategies are examined in Section~\ref{sec:fine_grained}.

\begin{figure}[!t]
\centering
\includegraphics[width=0.48\textwidth]{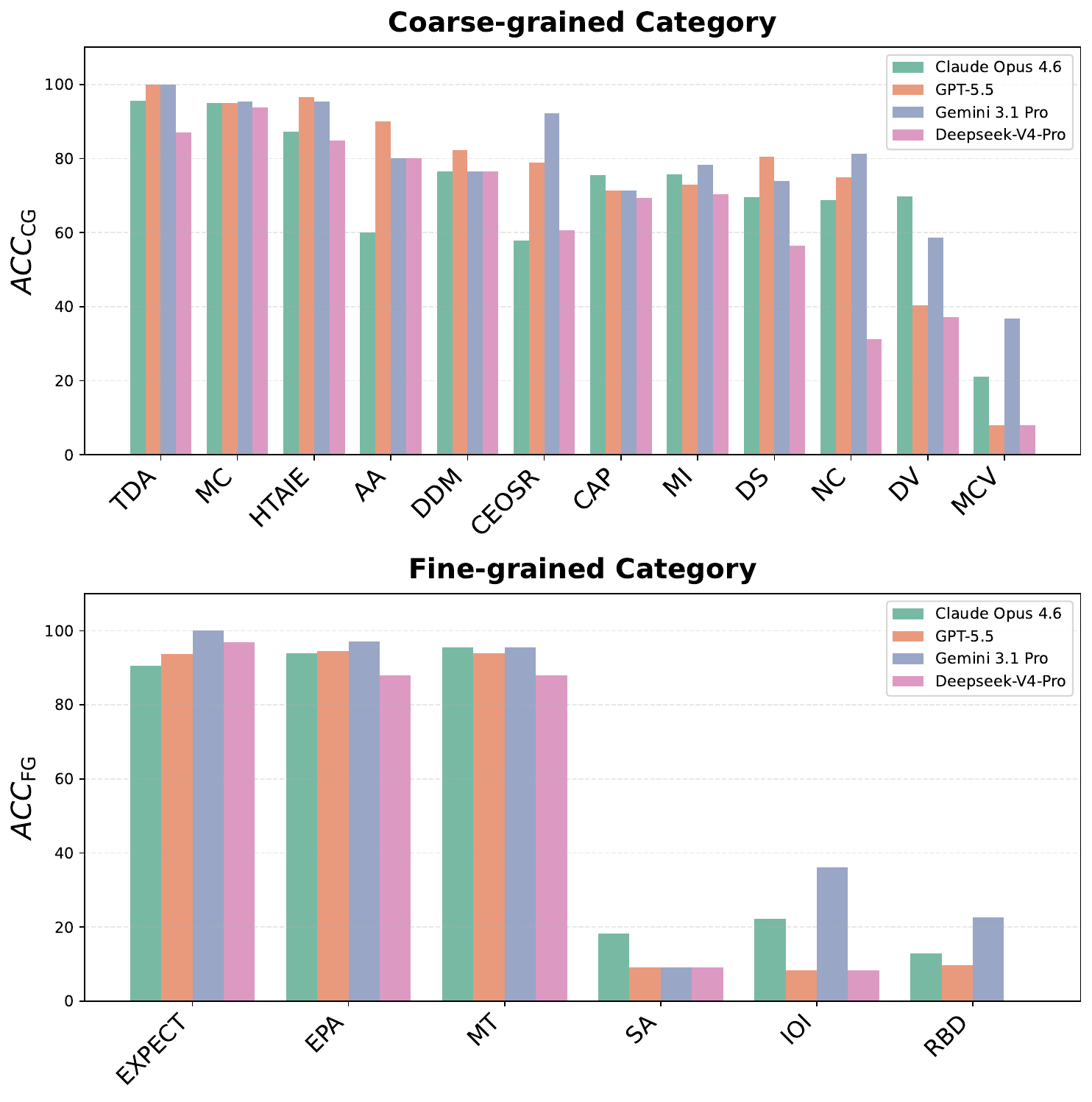}
\caption{Classification accuracy by category. Top: coarse-grained categories with more than 10 samples. Bottom: fine-grained categories with the highest and lowest average accuracy. Category abbreviations are defined in Table~\ref{tab:taxonomy}.}
\label{fig:acc_by_category}
\end{figure}

\subsection{Main Results}
\label{sec:main_results}

Table~\ref{tab:main_results} presents the main results across 14 LLMs and reveals persistent challenges in both formulation tasks. For statistical problem classification, Gemini 3.1 Pro achieves the highest fine-grained accuracy at 72.0, followed by Claude Opus 4.6 at 69.5 and Gemini 3 Flash at 68.4. The strongest open-source models, Kimi-K2.5, DeepSeek-V4-Flash, and Qwen3.5-397B-A17B, perform similarly on this task and trail the best closed-source model by about 7 points. Across all models, coarse-grained accuracy is consistently higher than fine-grained accuracy, indicating that models can often capture the broad analytical intent but still struggle to distinguish closely related subtypes.

Model rankings differ for variable identification and role assignment. Claude Opus 4.6 achieves the highest JCV at 63.2 and also leads on RV and VRI, while Gemini 3 Flash obtains the highest PV. Meanwhile, performance of models can also diverge substantially between the two tasks. Both GPT-5.4 and GPT-5.5 exceed 65 on $\mathrm{ACC}_{\mathrm{FG}}$ but remain near 52 on JCV. Even within the same model family, the relative ordering varies across metrics. Gemini 3 Flash surpasses Gemini 3.1 Pro on all four variable-related metrics, while DeepSeek-V4-Flash outperforms DeepSeek-V4-Pro on RV and VRI but trails it on JCV and PV. Taken together, these results show that no model consistently leads across both formulation tasks, leaving statistical problem formulation beyond the reliable reach of current LLMs.

\subsection{Fine-Grained Analysis}
\label{sec:fine_grained}

\paragraph{Per-category performance.}
As shown in Figure~\ref{fig:acc_by_category}, performance varies substantially across problem categories. At the coarse level, categories with explicit analytical goals, such as mathematical calculation and hypothesis testing, are easier, whereas categories that require distinguishing closely related analytical intents, such as model comparison and model interpretation, are more difficult. This contrast becomes more pronounced at the fine-grained level. Models perform well on categories with clear operational cues but struggle when category boundaries depend on subtle differences in the user request or data representation.

\begin{table}[t]
\centering
\small
\begin{tabular}{ll ccc}
\toprule
\textbf{Source} & \textbf{Model} & $\mathbf{ACC}_{\mathbf{FG}}$ & \textbf{JCV} & \textbf{VRI} \\
\midrule
\multirow{4}{*}{Book}
 & Claude Opus 4.6 & 78.5 & 51.0 & 56.9 \\
 & Gemini 3.1 Pro & 83.8 & 48.1 & 50.9 \\
 & GPT-5.5 & 84.1 & 33.6 & 47.5 \\
 & DeepSeek-V4-Pro & 74.1 & 48.6 & 50.0 \\
\midrule
\multirow{4}{*}{Case}
 & Claude Opus 4.6 & 54.7 & 83.2 & 81.3 \\
 & Gemini 3.1 Pro & 52.6 & 83.6 & 83.1 \\
 & GPT-5.5 & 39.6 & 82.4 & 83.3 \\
 & DeepSeek-V4-Pro & 34.6 & 82.5 & 77.2 \\
\bottomrule
\end{tabular}
\caption{Model performance across data sources.}
\label{tab:data_source_impact}
\end{table}

\paragraph{Data source impact.}
We evaluate representative models separately on the two data sources and observe a clear asymmetry, as shown in Table~\ref{tab:data_source_impact}. On textbook samples, $\mathrm{ACC}_{\mathrm{FG}}$ ranges from 74.1 to 84.1, while JCV ranges from 33.6 to 51.0. On case-library samples, the pattern reverses, with JCV between 82.4 and 83.6 but $\mathrm{ACC}_{\mathrm{FG}}$ between 34.6 and 54.7. This contrast reflects the characteristics of the two subsets. Textbook problems align better with standard statistical categories because they are typically designed around clearly defined problem types, making classification easier, but their compact data presentations demand careful variable extraction. Case-library problems instead provide structured datasets whose columns correspond directly to variables, making extraction easier, while their proximity to real-world analytical scenarios makes the problem types harder to classify.

\begin{table}[t]
\centering
\small
\begin{tabular}{l ccc ccc ccc}
\toprule
\textbf{Model} & \multicolumn{3}{c}{$\mathbf{ACC}_{\mathbf{FG}}$} & \multicolumn{3}{c}{\textbf{JCV}} & \multicolumn{3}{c}{\textbf{VRI}} \\
\midrule
\multirow{1}{*}{Claude Opus 4.6} & \multicolumn{3}{c}{69.5} & \multicolumn{3}{c}{\textbf{63.2}} & \multicolumn{3}{c}{\textbf{66.2}} \\
 \ \textit{w/} Category Def. & \multicolumn{3}{c}{\textcolor{green!60!black}{+2.8}} & \multicolumn{3}{c}{\textcolor{red!70!black}{-2.0}} & \multicolumn{3}{c}{\textcolor{green!60!black}{+0.6}} \\
\midrule
\multirow{1}{*}{GPT-5.5}
 & \multicolumn{3}{c}{67.2} & \multicolumn{3}{c}{52.1} & \multicolumn{3}{c}{61.1} \\
 \ \textit{w/} Category Def. & \multicolumn{3}{c}{\textcolor{green!60!black}{+1.9}} & \multicolumn{3}{c}{\textcolor{green!60!black}{+0.8}} & \multicolumn{3}{c}{+0.0} \\
\midrule
\multirow{1}{*}{Gemini 3.1 Pro}
 & \multicolumn{3}{c}{\textbf{72.0}} & \multicolumn{3}{c}{61.5} & \multicolumn{3}{c}{63.1} \\
 \ \textit{w/} Category Def. & \multicolumn{3}{c}{\textcolor{green!60!black}{+3.8}} & \multicolumn{3}{c}{\textcolor{red!70!black}{-0.7}} & \multicolumn{3}{c}{\textcolor{red!70!black}{-0.9}} \\
\bottomrule
\end{tabular}
\caption{Impact of prompting strategies. Zero-shot reports absolute scores, and other rows report \textcolor{green!60!black}{positive} or \textcolor{red!70!black}{negative} changes from zero-shot.}
\label{tab:prompting_strategies}
\end{table}

\begin{figure}[t]
\centering
\includegraphics[width=1.0\linewidth]{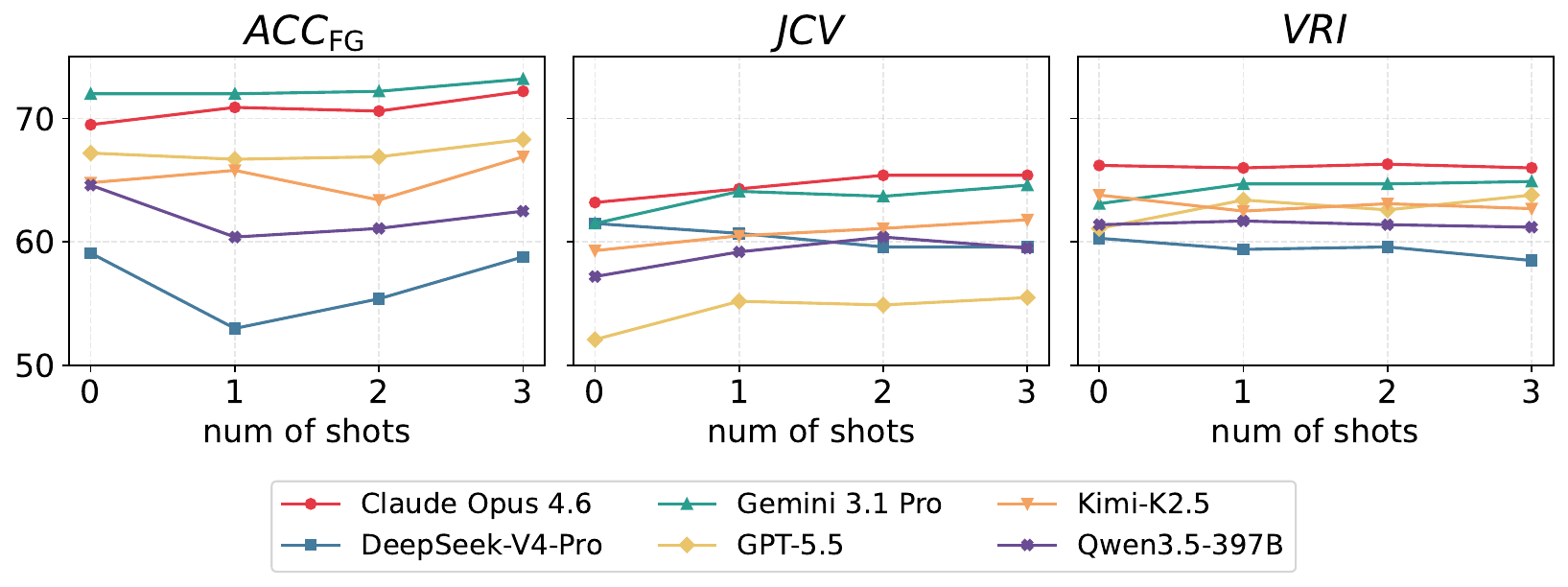}
\caption{Few-shot performance across prompting settings and metrics.}
\label{fig:prompting_strategy}
\end{figure}

\paragraph{Prompting with category definitions.}
Prompting strategies are known to affect LLM performance, and we examine their impact on formulation. As shown in Table~\ref{tab:prompting_strategies}, providing category definitions improves fine-grained classification for all three models by 1.9 to 3.8 points, indicating that explicit taxonomy information helps models distinguish closely related labels. Its effects on JCV and VRI are smaller and mixed, with most changes within one point. This suggests that explicit definitions can clarify category distinctions, but benefit to variable identification and role assignment is inconsistent and may be offset by the additional context.

\begin{table*}[t]
\centering
\small
\setlength{\tabcolsep}{4pt}
\begin{tabular}{@{}p{2.35cm} p{3.3cm} p{5.0cm} p{4.35cm}@{}}
\toprule
\textbf{Error case} & \textbf{Input} & \textbf{Prediction} & \textbf{Truth} \\
\midrule
GPT-5.4\par Variables
&
Among 41 patients, 26 had tumors on the same side as phone use. Compute the standardized evidence score for departure from an even-split benchmark.
&
\textbf{Category:}\par
Proportion Test

\vspace{0.4em}
\textbf{Variables:}\par
\texttt{same\_side\_count},\par
\texttt{total\_patients\_count},\par
\texttt{hypothesized\_proportion},\par
{\color{red!70!black}
\texttt{observed\_same\_side\_proportion}
}
&
\textbf{Category:}\par
Proportion Test

\vspace{0.4em}
\textbf{Variables:}\par
\texttt{same\_side\_count},\par
\texttt{total\_patients\_count},\par
\texttt{hypothesized\_proportion}
\\
\midrule
DeepSeek-V4-Pro\par Category
&
What is the distribution of short-term rental reviews (\texttt{vol\_num}) like?
&
\textbf{Category:}\par
\textcolor{red!70!black}{Distribution of Continuous Variables}

\vspace{0.4em}
\textbf{Variables:}\par
\texttt{vol\_num}
&
\textbf{Category:}\par
Distribution of Discrete Variables

\vspace{0.4em}
\textbf{Variables:}\par
\texttt{vol\_num}
\\
\bottomrule
\end{tabular}
\caption{Representative errors, with irrelevant fields omitted. \textcolor{red!70!black}{Red text} marks incorrect parts of model outputs.}
\label{tab:typical_errors}
\end{table*}

\paragraph{Few-shot prompting.}
Figure~\ref{fig:prompting_strategy} shows the effect of few-shot examples across six models. For $\mathrm{ACC}_{\mathrm{FG}}$, models with stronger zero-shot performance improve slightly as the number of shots increases, whereas lower-performing models generally drop at one shot before partially recovering. For VRI, Gemini 3.1 Pro and GPT-5.5 improve, while the remaining models stay nearly flat or decline. JCV improves modestly for most models, with DeepSeek-V4-Pro as the main exception. This pattern suggests that examples may help models align their variable outputs with the expected representation but do not resolve the conceptual distinctions required for category classification and role assignment. Overall, few-shot examples do not substantially improve statistical problem formulation.

\begin{figure}[t]
\centering
\includegraphics[width=0.48\textwidth]{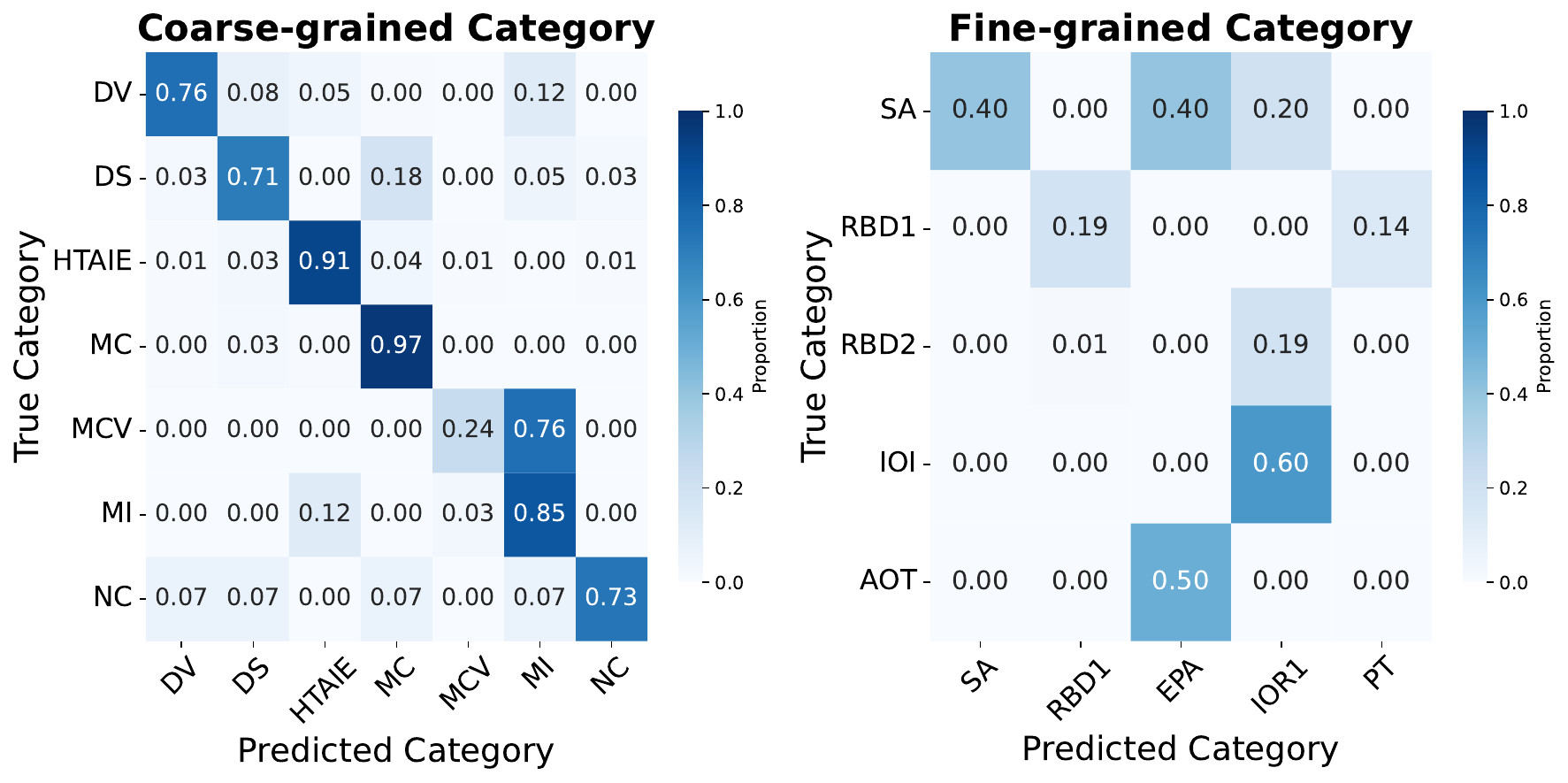}
\caption{Confusion matrices for hierarchical classification, focusing on frequently confused categories (we only present the \textbf{key parts} of matrices). Values are normalized by true labels.}
\label{fig:error_type}
\end{figure}

\subsection{Error Analysis}
\label{sec:case_study}

The confusion matrices in Figure~\ref{fig:error_type} show that models often confuse categories that share surface cues but differ in analytical intent. For example, model comparison is frequently predicted as model interpretation. At the fine-grained level, many errors remain within the correct first-level category, suggesting that models may recover the broad analytical intent yet fail to distinguish closely related subtypes.

\paragraph{Representative cases.}
Table~\ref{tab:typical_errors} presents two cases that illustrate how these confusions arise in individual samples. In the first case, GPT-5.4 correctly identifies the problem type but extracts a redundant derived quantity, whereas the reference label records only the data objects required for the calculation. This over-specification is consistent with the tendency toward redundant variable extraction observed in the main results. In the second case, DeepSeek-V4-Pro correctly identifies the relevant variable but treats it as continuous rather than discrete, which leads to an incorrect fine-grained visualization category. These cases confirm that statistical formulation demands joint reasoning across data representation and analytical intent, where a single misjudgment can propagate through the entire formulation chain.

\section{Conclusion}
\label{sec:conclusion}

We study statistical problem formulation as an evaluation target for LLMs and decompose it into statistical problem classification and variable identification with role assignment. To support this evaluation, we construct \StatFormBench{}, a benchmark of 1,013 samples drawn from five cross-domain statistics textbooks and a data science case library. The benchmark covers 20 coarse-grained and 85 fine-grained problem categories, and presents problems in scenario-style language that reflects how practitioners describe analytical needs.

Experiments on 14 LLMs show that no model performs consistently best across the two formulation tasks. The highest zero-shot fine-grained classification accuracy is 72.0, while the highest variable set overlap is 63.2, with the two results achieved by different models. Fine-grained analyses further show that performance of LLMs varies substantially across problem categories, while prompting strategies yield only limited and inconsistent gains. Taken together, these findings indicate that statistical problem formulation remains beyond the reliable reach of current models. Bridging this gap may require tighter integration of domain knowledge, statistical reasoning, and data-aware inference. We hope \StatFormBench{} helps clarify where current models fall short and supports the development of more capable statistical formulation assistants.

\section*{Limitations}
\label{sec:limitations}

Our benchmark has several limitations. First, the samples are derived from solved textbook exercises and a case library rather than real-world statistical consultations. This reflects a trade-off between realism and data accessibility, since real consultations often contain sensitive information and are subject to confidentiality constraints. We use constrained scenario rewriting to bring these curated sources closer to real consultations, but the resulting samples may still not fully capture the ambiguity and underspecification of real consulting. Second, the category distribution is imbalanced, with a substantial portion of samples concentrated in a few common types, such as data visualization and hypothesis testing. As a result, performance estimates for low-frequency categories may be less reliable. Third, the textbook subset is originally in English, whereas the case-library subset is originally in Chinese and translated for evaluation. This bilingual origin may introduce subtle cross-lingual effects that are not explicitly considered in the current evaluation.

\section*{Ethical Statements}
This work does not involve human subjects, and no personally identifiable information appears in the dataset. The textbook exercises are drawn from published educational materials. The case-library data were pre-anonymized by the data provider before we received them, with all personal names, organizational identifiers, and other potentially identifying attributes removed or replaced with placeholders. We manually verified a random subset and confirmed the absence of residual identifiable content.

\section*{Acknowledgements}
This work is supported by Shenzhen Science and Technology Program (Grant No. AI2026019).
We thank GouXiongHui for providing the case-library data used to construct the benchmark. We also thank Shuang Chen, Jiepeng Lai, Zikai Lin, Zichong Wang, and Ruitong Zhang for their careful review and verification of the textbook annotations.

\FloatBarrier

\bibliography{custom}

\appendix
\section{Data Processing and Annotation Details}
\label{sec:appendix_data}

We provide implementation details for the pipeline stages described in Section~\ref{sec:pipeline}. Prompt templates for each LLM-powered step are collected in Appendix~\ref{sec:appendix_prompts}.

\subsection{Textbook Processing}

\paragraph{Sample extraction.}
We use MinerU as the initial PDF parser, which emits structured outputs for body text, equations, tables, and figures. Annotators then manually correct the parser output for each problem, including fixing OCR errors in equations, replacing rasterized math with LaTeX, and transcribing tables that the parser misses. For each problem they produce a JSON entry containing the question text, reference answer, data (tabular, symbolic, or figure), and metadata (problem index, chapter/section identifiers, page range). The structured triple $(Q^o, A^o, D)$ is directly extracted from this entry. Worked examples and end-of-section exercises share the same protocol, except that for the latter the reference answer typically lives in a separate solutions appendix. All entries undergo a final proofreading pass before entering the next stage.

\paragraph{Sub-question splitting.}
Once extraction is complete, an LLM (\texttt{gpt-4o}) identifies the shared scenario in each problem and decomposes it into a background $\tilde{B}$ together with individual $(\tilde{Q}, \tilde{A})$ pairs. A deterministic flattening step then turns each pair into an independent sample that inherits $\tilde{B}$ and $D$, ensuring that every downstream sample carries exactly one analytical question.

\paragraph{Quality filtering.}
Each split sample is then evaluated by an LLM (\texttt{gpt-5.4}) against a set of exclusion criteria, producing a binary retain-or-discard label. Discarded samples include problems too basic to test statistical reasoning (such as looking up a single datum or pure concept recall without computation, though graphing and numerical calculation problems are retained), problems with incomplete information (e.g., those depending on an earlier problem's result), and problems whose reference answer is empty.

\paragraph{Initial label construction.}
For retained samples, the annotator LLM (\texttt{claude-opus-4-6}) produces initial labels in two stages. In the category labeling stage, the model selects a category in \texttt{L1\textbackslash L2} format from the current taxonomy, or proposes a new category when none fits. In the subsequent variable extraction stage, it produces variable labels in the schema $\{\textit{id}, \textit{value}, \textit{class}, \textit{role}, \textit{description}\}$, grounding each variable in the solving steps of $\tilde{A}$. These model-generated labels serve only as initial candidates and are subsequently verified by human reviewers as described in Section~\ref{sec:human_verification}.

\paragraph{Scenario rewriting.}
Finally, \texttt{gpt-5} rewrites $(\tilde{B},\tilde{Q})$ into $(B,Q)$, rephrasing statistical terminology into intuitive language, switching to a first-person collective voice, and preserving all numerical values. A self-evaluation prompt then checks whether the rewrite drifts from the original intent and triggers a one-shot correction when it does.

\subsection{Case-Library Adaptations}
\label{sec:appendix_case_library}

Compared with the textbook subset, the case-library subset differs only in the sample extraction stage, as the raw materials arrive in a fundamentally different form.

\paragraph{Project selection and data simulation.}
The source collection of cases from GouXiongHui contains 142 data-analysis projects with reports, code, and data in heterogeneous formats. We use only projects that include a complete analysis report and at least one data file that can be converted into a structured tabular representation. Image, audio, and other unstructured media files are then excluded, while the remaining tabular, serialized, and text data are converted into \texttt{pandas.DataFrame} objects. Each data frame is then summarized into a data dictionary containing the table shape, column names, data types, per-column descriptive statistics, missing-value counts, and a five-row preview. To protect the data provider's copyright, we further replace the original five-row preview with simulated rows generated by GPT-5 under constraints that preserve each field's data type, value range, and distributional characteristics.

\paragraph{Question induction and annotation.}
For each retained project, the HTML analysis report and data dictionary are provided to \texttt{gpt-4o-mini}, which produces a case overview and a set of independent analysis questions. Each question is paired with a report-grounded answer and the variables needed for analysis. A second \texttt{gpt-4o-mini} call checks whether the questions are of moderate difficulty, mutually independent, and supported by the report. Outputs that fail these criteria are revised using the model's feedback for up to five rounds. After a question passes this check, separate \texttt{gpt-4o} calls produce its initial category and variable annotations. This process directly produces independent $(\tilde{B}, \tilde{Q}, \tilde{A})$ tuples, making sub-question splitting unnecessary for the case-library subset.

\section{Statistical Problem Taxonomy}
\label{sec:appendix_taxonomy}

The taxonomy contains $20$ first-level and $85$ second-level categories, designed by jointly considering canonical statistical problem types from standard practice and problem types that surface in our collected samples. Table~\ref{tab:taxonomy} lists all categories with brief descriptions derived from the operational definitions used in annotation. We do not claim that the taxonomy exhausts every statistical problem type, and we expect it to grow as additional data sources are incorporated.

\begin{table*}[p]
\centering
\scriptsize
\setlength{\tabcolsep}{2.5pt}
\renewcommand{\arraystretch}{1.00}
\begin{minipage}[t]{0.48\textwidth}
\vspace{0pt}
\centering
\begin{tabular}{@{}p{1.5cm}lp{1.7cm}lp{2.5cm}@{}}
\toprule
\textbf{L1 Category} & \textbf{Abbr} & \textbf{L2 Category} & \textbf{Abbr} & \textbf{Description} \\
\midrule
\textbf{Data Preproc.} & \textbf{DP} & Missing Value Handling & MVH & Impute or delete missing data \\
& & Outlier Handling & OH & Detect and treat anomalies \\
& & Standardization & STANDA & Rescale to comparable units \\
& & Binning & BINNIN & Discretize into intervals \\
& & Transformation & TRANSF & Log, Box-Cox, unit conversion \\
\addlinespace[2pt]
\textbf{Math.\ Calc.} & \textbf{MC} & Prob.\ Space Related & PSR & Count configurations or states \\
& & Event Prob.\ \& Independence & EPA & Probability rules, independence \\
& & Expectation & EXPECT & Expected values, long-run averages \\
& & Min.\ Sample Size for Target Prob. & MSS & Min $n$ for a probability target \\
& & Distribution Derivation & DD & Derive probability distributions \\
& & Other & OTHER & Misc.\ probability calculations \\
\addlinespace[2pt]
\textbf{Numer.\ Comp.} & \textbf{NC} & Model Param.\ Estimation & OMP & Analytic or numeric estimation \\
& & Sampling from Spec.\ Distrib. & SFA & Generate random draws \\
\addlinespace[2pt]
\textbf{Descr.\ Stat.} & \textbf{DS} & Central Tendency & MOC & Mean, median, mode \\
& & Dispersion & MOD1 & Variance, SD, CV, IQR \\
& & Distribution Range & MOD2 & Range, min, max \\
& & Frequency \& Proportion & MOF & Counts, proportions, risk ratios \\
& & Quantiles / Percentiles & QUANTI & Values at specified ranks \\
& & Network Graph Indicators & NGI & Degree, centrality, density \\
& & Stat.\ Property Comparison & AAC & Compare estimator properties \\
& & Other (e.g., reliability) & OR & Cronbach's $\alpha$, consistency \\
\addlinespace[2pt]
\textbf{Dist.\ Model.} & \textbf{DDM} & Model Building & MB & Choose distribution family \\
& & MLE & MLE & Maximum likelihood fitting \\
& & Bayesian & BAYESI & Posterior from prior and data \\
\addlinespace[2pt]
\textbf{Data Visual.} & \textbf{DV} & Discrete Var.\ Distrib. & DOD & Bar, pie charts \\
& & Continuous Var.\ Distrib. & DOC & Histogram, density, box plot \\
& & Discrete \& Continuous & RBD1 & Grouped box / density plots \\
& & Continuous \& Continuous & RBC & Scatter plots, fitted curves \\
& & Discrete \& Discrete & RBD2 & Grouped bar, mosaic plots \\
& & Multi-Var.\ Comparison & CCO & Heatmap, radar, parallel coords \\
& & Trends Over Time & TOT & Line, time-series, area charts \\
& & Spatial \& Geographic & SAG & Choropleth, spatial scatter \\
& & Text \& Symbolic & TAS & Text / symbol visualization \\
& & Stem-and-Leaf Plot & SP & Distribution with raw digits \\
& & Other Stat.\ Graphics & OSG & Specialized statistical plots \\
\addlinespace[2pt]
\textbf{Association} & \textbf{AA} & Correlation Coeff.\ (1-D) & CC & Pearson, Spearman, Kendall \\
& & CCA (Multidimensional) & CCA & Canonical correlation \\
& & Contingency Table & CT & Joint categorical distribution \\
\addlinespace[2pt]
\textbf{Clustering} & \textbf{CLU} & Clustering & CLUSTE & Unsupervised grouping \\
\addlinespace[2pt]
\textbf{Dim.\ Reduct.} & \textbf{DR} & Dimensionality Reduction & DR & PCA, t-SNE, UMAP \\
\bottomrule
\end{tabular}
\end{minipage}
\hfill
\begin{minipage}[t]{0.48\textwidth}
\vspace{0pt}
\centering
\begin{tabular}{@{}p{1.5cm}lp{1.7cm}lp{2.5cm}@{}}
\toprule
\textbf{L1 Category} & \textbf{Abbr} & \textbf{L2 Category} & \textbf{Abbr} & \textbf{Description} \\
\midrule
\textbf{Hyp.\ Testing}$^\dagger$ & \textbf{HTAIE} & Independence Test & IT & $\chi^2$, Fisher exact \\
& & Mean Test & MT & $t$-test, paired, two-sample \\
& & Variance Test & VT & Compare population variances \\
& & ANOVA & ANOVA & Group variation decomposition \\
& & Distribution Test & DT & KS, Shapiro-Wilk \\
& & Proportion Test & PT & Compare group proportions \\
& & Likelihood Ratio Test & LRT1 & Nested model comparison \\
& & Sequential Test & ST & Sequential testing (SPRT) \\
& & Randomness & RANDOM & Runs test, randomness test \\
& & Regression Param.\ Test & RMP & Coefficient significance test \\
& & Test Properties & AOT & Power, size, Type I/II error \\
\addlinespace[2pt]
\textbf{Model Comp.} & \textbf{MCV} & Criterion-Based (AIC, BIC) & CCB & Information criteria comparison \\
& & Cross-Validation & CROSS- & Data-splitting generalization \\
& & Likelihood Ratio Test & LRT2 & Nested model selection by LR \\
& & ROC / AUC & RCC & Classifier discrimination \\
& & Important Var.\ Identif. & IOI & Stepwise, LASSO selection \\
\addlinespace[2pt]
\textbf{Model Diag.} & \textbf{MD} & Residual Analysis & RA & Residual patterns, influence \\
& & Goodness-of-Fit & GA & $R^2$, adjusted $R^2$ \\
\addlinespace[2pt]
\textbf{Model Interp.} & \textbf{MI} & Coeff.\ Signif.\ \& Direction & IOR1 & Interpret coefficient sign \\
& & Effect Size & IOR2 & Magnitude of predictor effects \\
& & Classif./Occur.\ Prob. & IOF & Interpret OR, HR \\
& & Ordered Outcome Mech. & IOM & Ordinal model interpretation \\
& & Variable Importance & IOV & SHAP, standardized coefficients \\
& & Group Heterogeneity & IOG & Subgroup effect comparison \\
\addlinespace[2pt]
\textbf{Classif./Pred.} & \textbf{CAP} & Continuous Value Pred. & CVP & Predict numerical outcomes \\
& & Categorical Classif. & CVC & Predict unordered classes \\
& & Ordered Multi-Class & MCO & Predict ordinal outcomes \\
& & Quantile Prediction & QP & Conditional quantile prediction \\
& & Survival Analysis & SA1 & Time-to-event (KM, Cox) \\
\addlinespace[2pt]
\textbf{Time Series} & \textbf{TSA} & Lag Correlation & LCA & ACF, PACF analysis \\
& & Stationarity & SA2 & Unit root tests (ADF, KPSS) \\
& & Seasonality & SA3 & Periodic pattern detection \\
& & Factor Decomposition & DFD & Trend, seasonal, cycle \\
& & Volatility & VA & ARCH/GARCH modeling \\
\addlinespace[2pt]
\textbf{Text Data} & \textbf{TDA} & Word Frequency & WFS & Token count and ranking \\
& & Topic Modeling & TM & Latent topics (LDA) \\
& & Sentiment Analysis & SA4 & Text polarity classification \\
\addlinespace[2pt]
\textbf{Network Data} & \textbf{NDA} & Community Detection & CD & Dense node group identification \\
& & Link Prediction & LP & Missing / future edge prediction \\
\addlinespace[2pt]
\textbf{Exper.\ Design} & \textbf{ED} & Plan Design & EPD & Design experiments / sampling \\
& & Plan Evaluation & EPE & Evaluate sampling quality \\
\addlinespace[2pt]
\textbf{Causal Inf.} & \textbf{CI} & Observational Data & OD & PSM, IV, DiD estimation \\
& & Experimental Data & ED & RCT, A/B test effects \\
\addlinespace[2pt]
\textbf{Critical Eval.} & \textbf{CEOSR} & Stat.\ Fallacies \& Biases & IOS & Detect statistical misuse \\
& & Result Reliability & EOR & Assess result reliability \\
\bottomrule
\end{tabular}
\end{minipage}
\caption{Complete statistical problem taxonomy ($20$ first-level, $85$ second-level categories). $^\dagger$\,L2 categories under Hypothesis Testing also cover the corresponding interval-estimation tasks.}
\label{tab:taxonomy}
\end{table*}

\section{Prompt Templates}
\label{sec:appendix_prompts}

We present the prompt templates used in each LLM-powered pipeline step (Section~\ref{sec:pipeline}), listed in pipeline order.

\begin{figure*}[t]
\begin{tcolorbox}[promptbox, title=Quality Filtering Prompt (\texttt{gpt-5.4})]
You are a professional statistics instructor. Judge whether the following exercise is suitable for testing statistical proficiency.

\textbf{Input:} \texttt{\{background, data, question, answer\}}

Judge the exercise as \textbf{unsuitable} (\texttt{judge}=0) if any of the following conditions applies:
\begin{itemize}[nosep,leftmargin=*]
\item \textit{Overly basic:} the exercise requires only a single data lookup or pure concept recall without computation. Retain graphing and numerical calculation problems.
\item \textit{Incomplete information:} the description is ambiguous, or the question depends on a preceding problem's result that is not provided.
\item \textit{Empty answer:} the \texttt{answer} field is blank.
\end{itemize}
Note: tables referenced in \texttt{question} may appear as LaTeX code in \texttt{data}. This does not count as missing information.

\textbf{Output:} \texttt{\{"judge": 1/0, "explanation": "..."\}}
\end{tcolorbox}
\end{figure*}

\begin{figure*}[t]
\begin{tcolorbox}[promptbox, title=Category Annotation Prompt (\texttt{claude-opus-4-6})]
You are a strict statistical problem category annotator. Given a sample's \texttt{background}, \texttt{data}, and \texttt{question}, select the most appropriate category label from the statistical problem taxonomy below.

\textbf{Requirements:}
\begin{enumerate}[nosep,leftmargin=*]
\item If a fitting L2 category exists, write it strictly as ``L1\textbackslash L2'' in \texttt{suggested\_category}.
\item If no L2 category fits, fill \texttt{proposed\_new\_category} instead and leave \texttt{suggested\_category} empty.
\item \texttt{reason} must explain the choice or why a new category is needed.
\end{enumerate}

\textbf{Statistical Problem Taxonomy} (20 L1 and 85 L2 categories, abridged):\\[2pt]
\textbullet~\textit{Data Preprocessing}: [Missing Value Handling, Outlier Handling, Standardization, Binning, Transformation]\\
\textbullet~\textit{Mathematical Calculation}: [Probability Space, Event Probability \& Independence, Expectation, Min.\ Sample Size for Target Probability, Distribution Derivation, Other]\\
\textbullet~\textit{Hypothesis Testing \& Interval Estimation}: [Independence Test, Mean Test, Variance Test, ANOVA, Distribution Test, Proportion Test, Likelihood Ratio Test, Sequential Test, Randomness, Regression Parameter Test, Test Properties]\\
\quad\ldots\enspace(\textit{The remaining 17 L1 categories are listed in Appendix~\ref{sec:appendix_taxonomy}})\\[2pt]
Each L2 category is accompanied by a one-sentence definition in the actual prompt.

\textbf{Input:} \texttt{\{background, data, question\}}\\
\textbf{Output:} \texttt{\{"suggested\_category": "...", "proposed\_new\_category": "...", "reason": "..."\}}
\end{tcolorbox}
\end{figure*}

\begin{figure*}[t]
\begin{tcolorbox}[promptbox, title=Variable Extraction Prompt (\texttt{claude-opus-4-6})]
You are a strict statistical annotator. Extract the variable objects that the \texttt{answer} \textit{actually uses} in its solving process.

\textbf{Key requirements} (8 rules):\\
1.~Variables are data objects required to answer the research question, including raw columns, sample sizes, means, SDs, and contingency tables. \textbf{Exclude} model parameters and test statistics ($z$, $p$, $t$, $F$).\\
2.~Ground extraction in the answer's actual reasoning. Omit variables the answer never uses, and include variables that appear symbolically in \texttt{question} or \texttt{answer} even when \texttt{data} is empty.\\
3.~For grouped data, create one variable object per group.\\
4.~For per-variable summary statistics, store aggregates in the \texttt{value} field as JSON.\\
5.~Variable objects include raw columns, grouped statistics, contingency tables, sample sizes, and success counts. When the answer constructs a derived variable from multiple originals, retain all originals and omit the derived one.\\
6.~For ``Mathematical Calculation'' problems, probabilities, sample sizes, and other numerical objects used in the answer are also treated as variable objects.\\
7.~If the answer's solving process is missing or overly brief, infer the minimal variable set needed to reach the final answer. If no variables are needed, output empty JSON.\\
8.~Output must be valid JSON with variable \texttt{id} values as top-level keys. Do not include extra text.

\textbf{Fields per variable:}\\
\texttt{id}: A variable identifier.\\
\texttt{value}: Store a table column as an array, aggregated measures as JSON, and any other value in its original recorded form. When multiple numerical values are present, preserve their order in the source problem.\\
\texttt{class}: One of \texttt{numerical}, \texttt{categorical}, or \texttt{others}.\\
\texttt{role}: One of \texttt{predictor}, \texttt{response}, \texttt{both}, or \texttt{N/A}.\\
\texttt{description}: A brief description of the variable.\\
Roles: \texttt{predictor} denotes a predictor variable, \texttt{response} denotes a response variable, and \texttt{both} denotes an undirected association or a dual role. \texttt{N/A} indicates that no directional role is needed, as with identifiers, filters, unsupervised features, and single-variable distributions. Special-case rules for time-trend problems and variables serving dual construction roles are included in the full prompt.

\textbf{Input:} \texttt{\{background, data, question, answer\}}\\
\textbf{Output:} \texttt{\{"VAR": \{"id", "value", "class", "role", "description"\}, \ldots\}}
\end{tcolorbox}
\end{figure*}

\begin{figure*}[t]
\begin{tcolorbox}[promptbox, title=Scenario Rewriting Prompt (\texttt{gpt-5})]
You are a statistician. Rewrite the \texttt{background} and \texttt{question} of the given statistical problem to create a more realistic formulation challenge from a domain practitioner's perspective.

\textbf{Requirements:}
\begin{itemize}[nosep,leftmargin=*]
\item Use business language from the relevant domain. Do not use statistical terminology or explicitly state the target statistical problem category.
\item Switch to a first-person collective voice (``we'').
\item Preserve all numerical values, named entities, and table/figure labels.
\item The rewritten problem must still map to the same formalized statistical model.
\item The wording may be appropriately vague, and distracting context may be added.
\end{itemize}

The full statistical problem taxonomy and variable specification used in the two prompts above are also provided so the model can verify that the mapping is preserved. They are omitted here for brevity. Two worked examples demonstrating the rewriting style are included in the actual prompt.

\textbf{Input:} \texttt{\{background, data, question\}} + target statistical model\\
\textbf{Output:} \texttt{\{"background": "...", "question": "..."\}}
\end{tcolorbox}
\end{figure*}

\begin{figure*}[t]
\begin{tcolorbox}[promptbox, title=Rewrite Self-Evaluation Prompt (\texttt{gpt-5})]
You are an expert evaluator of problem reformulation. Given the original problem, its formal statistical model, the derivation reasoning, and the rewritten problem, check whether the rewrite meets all of the following standards:
\begin{enumerate}[nosep,leftmargin=*]
\item Numerical values in the original are preserved. No unauthorized new data is introduced.
\item The analytical direction of the original problem is maintained. No new problem is created.
\item Table and figure labels are unchanged.
\item The rewritten question does not explicitly state the target statistical problem category.
\item The rewritten problem still matches the given formal statistical model.
\end{enumerate}

\textbf{Input:} original problem, statistical model, derivation reasoning, rewritten problem\\
\textbf{Output:} \texttt{\{"pass": true/false, "feedback": "..."\}}
\end{tcolorbox}
\end{figure*}

\begin{figure}[t]
\begin{tcolorbox}[promptbox, title=Statistical Problem Formulation Prompt]
You are a professional statistician. Abstract the statistical formulation of the following problem.
The input format is:

\texttt{\{}
\texttt{    "background": "Relevant description of the problem context (which may also contain some data)",}
\texttt{    "data": "The data used, presented as a table in LaTeX code",}
\texttt{    "question": "The core problem"}
\texttt{\}}

The statistical formulation includes the following two parts:

1. Statistical problem category

\texttt{\{}\texttt{statistical problem taxonomy}
\texttt{\}}

2. Relevant variables and their roles

Key requirements:  
\begin{enumerate}[nosep, leftmargin=*]
\item Variables refer specifically to data objects collected for the research objective (e.g., sample size, mean, standard deviation), excluding model parameters, z-scores, p-values, t-values, or F-values.
\item Variable objects can be raw data columns, grouped statistics, contingency tables, sample sizes, success counts, or time variables.  
\item For "Mathematical Calculation" problems, treat numerical objects (probabilities, sample sizes) used in the answer as variables.  
\item If there is no explicit data table or if \texttt{data} is empty, variables must still be extracted when symbolic variables or statistics appear.
\item When the data have a grouping structure, split them into multiple variables by group.
\item If aggregated measures are provided for multiple variables, create a separate variable object for each, writing measures into the \texttt{value} field as JSON.  
\item If the information is insufficient, output an empty JSON object: \texttt{\{\}}.
\item Output must be valid JSON with variable IDs as top-level keys. Do not include extra text.
\end{enumerate}

Field specifications:  
\begin{itemize}[nosep, leftmargin=*]
\item Top-level key: Variable name (extracted or assigned).  
\item \texttt{id}: Must match top-level key.  
\item \texttt{value}: An array for table columns, JSON for aggregated measures, or the original form (e.g., a TeX table). When multiple numerical values are present, preserve their order in the source problem.
\item \texttt{class}: One of \texttt{numerical}, \texttt{categorical}, \texttt{others}.  
\item \texttt{role}: One of \texttt{predictor}, \texttt{response}, \texttt{both}, or \texttt{N/A}. Use \texttt{both} for an undirected association or a dual role. Use \texttt{N/A} when no directional role is needed, as with identifiers, filters, unsupervised methods, and distribution descriptions.
\item \texttt{description}: Brief description.
\end{itemize}

Special \texttt{role} cases:  
\begin{itemize}[nosep, leftmargin=*]
\item For "Trends Over Time" visualization, assign \texttt{predictor} to the time variable and \texttt{response} to the changing variable.
\item If a variable serves as a predictor and is also used to construct the response, assign \texttt{predictor}.
\item If a variable serves as a response and is also used to construct the predictor, assign \texttt{response}.
\end{itemize}

Note: Shallow correlation analyses typically use "Data Visualization." Deeper analyses of direction, magnitude, or significance use modeling and interpretation.
Time series forecasting uses "Continuous Value Prediction." "Time Series Analysis" focuses on interpretation.

Output format: JSON. Example:

\texttt{\{}
\texttt{  "category": "Relationship Between Continuous Variables",}
\texttt{  "variables": \{}
\texttt{    "height": \{}
\texttt{      "id": "height",}
\texttt{      "value": \{"$\mu$": 138, "$\sigma$": 7\},}
\texttt{      "class": "numerical",}
\texttt{      "role": "N/A",}
\texttt{      "description": "Heights of 10-year-old boys modeled as Normal($\mu$=138 cm, $\sigma$=7 cm)."}
\texttt{    \},}
\texttt{    "threshold": \{}
\texttt{      "id": "threshold",}
\texttt{      "value": 150,}
\texttt{      "class": "numerical",}
\texttt{      "role": "N/A",}
\texttt{      "description": "Cutoff value (150 cm) used in the z-score calculation z=(150-138)/7."}
\texttt{    \}}
\texttt{  \}}
\texttt{\}}
\end{tcolorbox}
\end{figure}

\FloatBarrier
\onecolumn
\section{Additional Dataset Statistics}
\label{sec:appendix_dataset_statistics}

\begin{figure}[htbp]
    \includegraphics[width=1\linewidth]{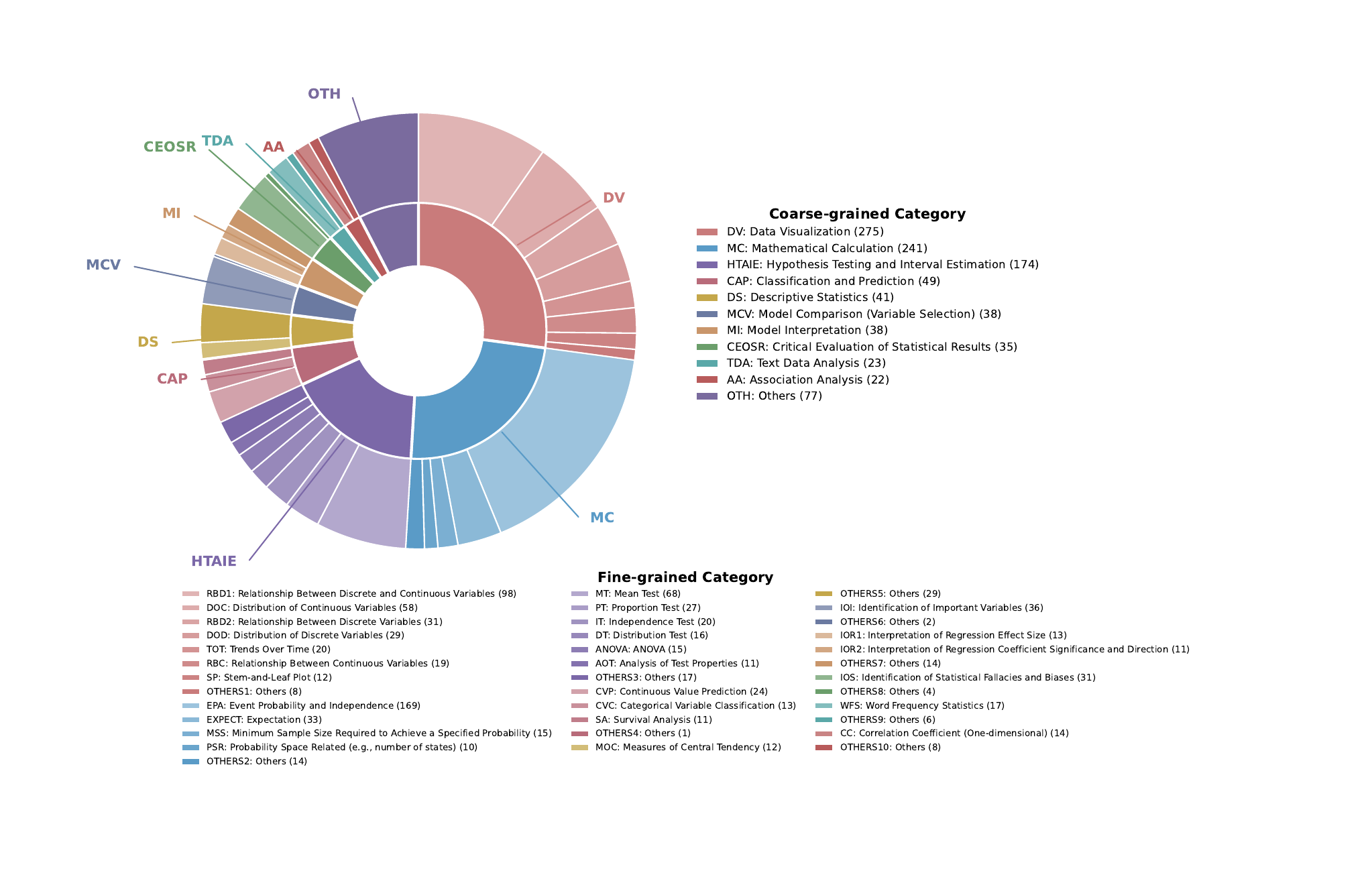}
    \caption{Category distribution at both taxonomy levels. The inner ring shows coarse-grained categories, with those containing no more than 20 samples grouped as Others. The outer ring shows the fine-grained composition of each displayed coarse-grained category.}
    \label{all_catoegory}
\end{figure}

Figure~\ref{all_catoegory} complements the dataset overview in Figure~\ref{fig:dataset_dist} by showing the fine-grained composition of the most frequent coarse-grained categories.

\FloatBarrier
\twocolumn
\section{Supplementary Results}
\label{sec:appendix_results}

\subsection{Exact-Value Matching Implementation and Validation}
\label{sec:appendix_exact_matching}

\paragraph{Matching procedure.}
As specified in the variable extraction prompt above, the \texttt{value} field records the complete contents of a variable under a standardized schema. We use this field as the matching key so that synonymous variable names or descriptions do not affect the score. The predicted and reference \texttt{value} fields are serialized, and a pair is matched only when the resulting strings are identical. No additional normalization or semantic matching is applied, which keeps the evaluation rule deterministic and reproducible. Each predicted and reference variable can appear in at most one matched pair, with unmatched predictions and reference variables counted as false positives and false negatives, respectively. Consequently, per-sample JCV equals one only when every predicted and reference variable can be paired through exact value equality.

\paragraph{Validation study.}
To quantify how often exact-value matching rejects semantically correct correspondences, we conduct a targeted validation study on a challenging subset. We define a matching-induced error as an output in which exact-value matching rejects at least one semantically correct variable correspondence. Such an error occurs when a correctly identified variable is treated as unmatched solely because the prediction uses a different variable decomposition or an equivalent value representation. For example, a model may represent a sample size and a mean as separate variables while the reference stores both within a single variable entry, or represent a percentage as \texttt{50} rather than \texttt{50\%}. The challenging subset is defined using three representative closed-source models: Claude Opus 4.6, GPT-5.5, and Gemini 3.1 Pro. It contains 387 of the 629 textbook samples on which all three models have per-sample JCV below one. We randomly sample 60 cases from this subset, and a PhD-level expert compares each model's predicted variable set with the reference label to determine whether any semantically correct correspondence is rejected.

\begin{table}[t]
\centering
\small
\begin{tabular}{@{}lc@{}}
\toprule
\textbf{Model} & \textbf{Matching-Induced Error Rate} \\
\midrule
Claude Opus 4.6 & 11.7\% (7/60) \\
GPT-5.5 & 10.0\% (6/60) \\
Gemini 3.1 Pro & 10.0\% (6/60) \\
\bottomrule
\end{tabular}
\caption{Matching-induced error rates on the validation subset.}
\label{tab:exact_matching_validation}
\end{table}

\paragraph{Results.}
As shown in Table~\ref{tab:exact_matching_validation}, only about 10\% of the sampled outputs for each model contain a matching-induced error, even in this challenging subset. The results suggest that although exact-value matching can underestimate variable-identification performance to some extent, it does not account for the substantially lower JCV scores in the main experiments.

\FloatBarrier
\onecolumn
\subsection{Prompting Strategies}

\begin{figure*}[htbp]
    \includegraphics[width=1\linewidth]{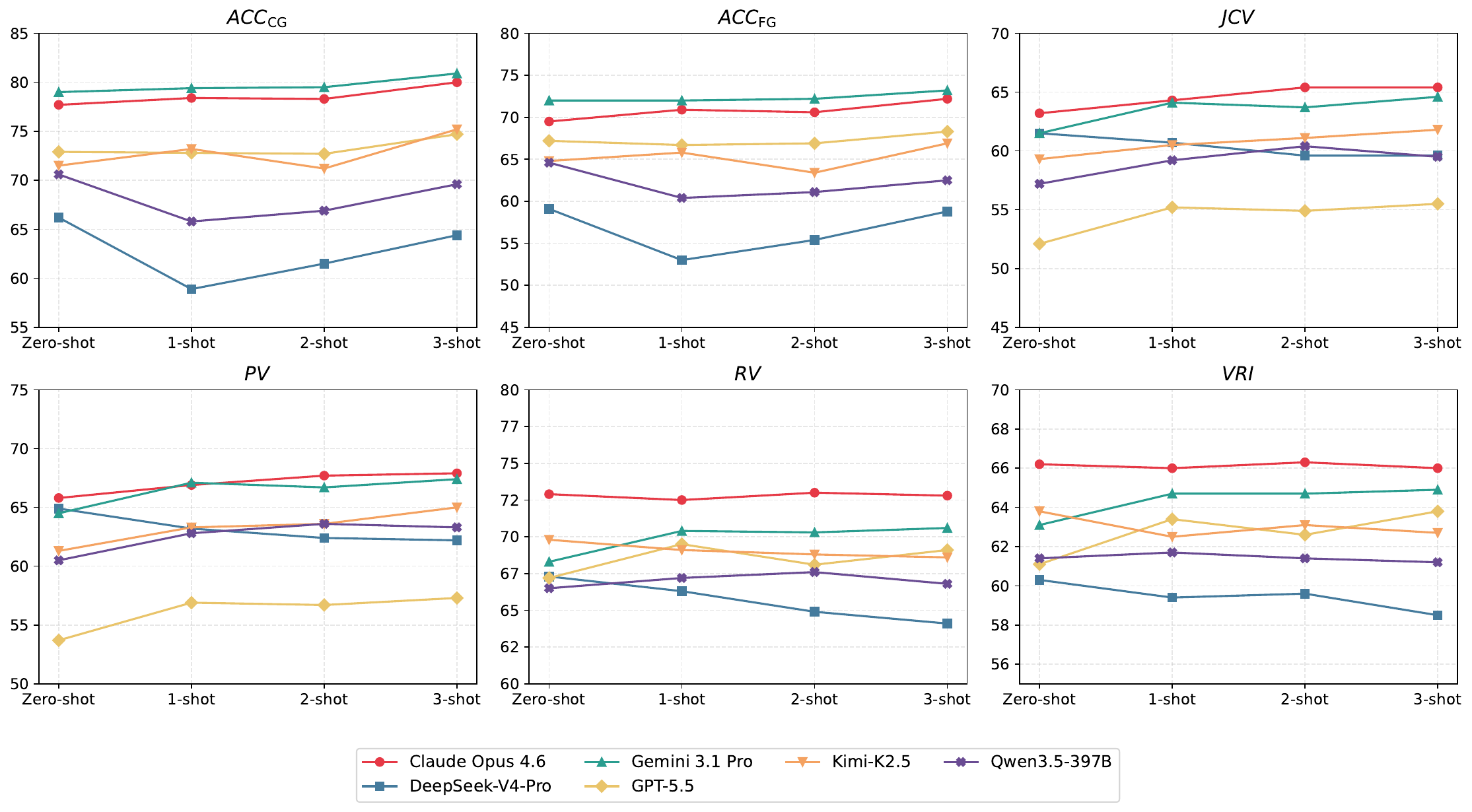}
    \centering
    \caption{Performance across prompting strategies on all six evaluation metrics.}

    \label{All_performance_metrics}
\end{figure*}

Figure~\ref{All_performance_metrics} supplements the few-shot analysis in Section~\ref{sec:fine_grained} by reporting performance on all six evaluation metrics under the same experimental setting. The complete comparison leads to the same conclusion that adding examples to the prompt does not yield consistent gains across the two evaluated tasks.

\FloatBarrier
\subsection{Full Comparisons by Data Source}
Table~\ref{tab:metrics_by_source_all} extends the source-based analysis to ten models and all six metrics under the same evaluation setting. Consistent with the pattern observed in Section~\ref{sec:fine_grained}, classification accuracy is higher on textbook samples, whereas variable-related scores are higher on case-library samples.

\begin{table*}[htbp]
\centering
\scriptsize
\setlength{\tabcolsep}{14pt}
\begin{tabular}{llcccccc}
\toprule
\multirow{2}*{\textbf{Source}} & \multirow{2}*{\textbf{Model}} & \multicolumn{2}{c}{\textbf{Classification}} & \multicolumn{4}{c}{\textbf{Variable Id.\ \& Role}} \\
\cmidrule(lr){3-4} \cmidrule(lr){5-8}
 & & $\mathrm{ACC}_{\mathrm{CG}}$ & $\mathrm{ACC}_{\mathrm{FG}}$ & JCV & PV & RV & VRI \\
\midrule
\multirow{10}{*}{\textbf{Book}} & Claude Opus 4.6 & 82.7 & 78.5 & 51.0 & 55.1 & 63.2 & 56.9 \\
 & Claude Sonnet 4.6 & 79.5 & 73.0 & 50.0 & 54.1 & 61.1 & 53.0 \\
 & GPT-5.5 & 88.6 & 84.1 & 33.6 & 36.7 & 54.3 & 47.5 \\
 & GPT-5.4 & 86.3 & 77.9 & 33.2 & 36.1 & 56.9 & 49.9 \\
 & Gemini 3.1 Pro & 88.7 & 83.8 & 48.1 & 52.8 & 57.2 & 50.9 \\
 & Gemini 3 Flash & 87.6 & 81.2 & 50.2 & 55.2 & 59.2 & 53.3 \\
 & Kimi-K2.5 & 84.9 & 79.2 & 46.6 & 50.1 & 59.6 & 52.3 \\
 & DeepSeek-V4-Pro & 80.1 & 74.1 & 48.6 & 54.1 & 55.7 & 50.0 \\
 & DeepSeek-V4-Flash & 87.0 & 81.9 & 37.3 & 39.9 & 60.3 & 54.9 \\
 & Qwen3.5-397B & 83.9 & 79.0 & 45.3 & 49.5 & 58.7 & 51.5 \\
\midrule
\multirow{10}{*}{\textbf{Case}} & Claude Opus 4.6 & 69.5 & 54.7 & 83.2 & 83.3 & 88.8 & 81.3 \\
 & Claude Sonnet 4.6 & 59.4 & 49.0 & 82.6 & 82.8 & 87.0 & 80.1 \\
 & GPT-5.5 & 47.1 & 39.6 & 82.4 & 81.5 & 88.4 & 83.3 \\
 & GPT-5.4 & 57.0 & 46.1 & 82.7 & 82.4 & 86.9 & 81.9 \\
 & Gemini 3.1 Pro & 63.0 & 52.6 & 83.6 & 83.7 & 86.5 & 83.1 \\
 & Gemini 3 Flash & 59.6 & 47.4 & 82.3 & 84.0 & 84.8 & 80.4 \\
 & Kimi-K2.5 & 49.5 & 41.1 & 80.1 & 79.7 & 86.6 & 82.6 \\
 & DeepSeek-V4-Pro & 43.5 & 34.6 & 82.5 & 82.6 & 86.3 & 77.2 \\
 & DeepSeek-V4-Flash & 45.3 & 36.7 & 82.6 & 82.6 & 88.2 & 82.2 \\
 & Qwen3.5-397B & 48.7 & 40.9 & 76.6 & 78.6 & 79.2 & 77.7 \\
\bottomrule
\end{tabular}
\caption{Model performance across six metrics for each data source.}
\label{tab:metrics_by_source_all}
\end{table*}

\newpage
\FloatBarrier
\subsection{All Results of Classification Accuracy by Category}

\begin{figure}[htbp]
\centering
\includegraphics[width=1\textwidth]{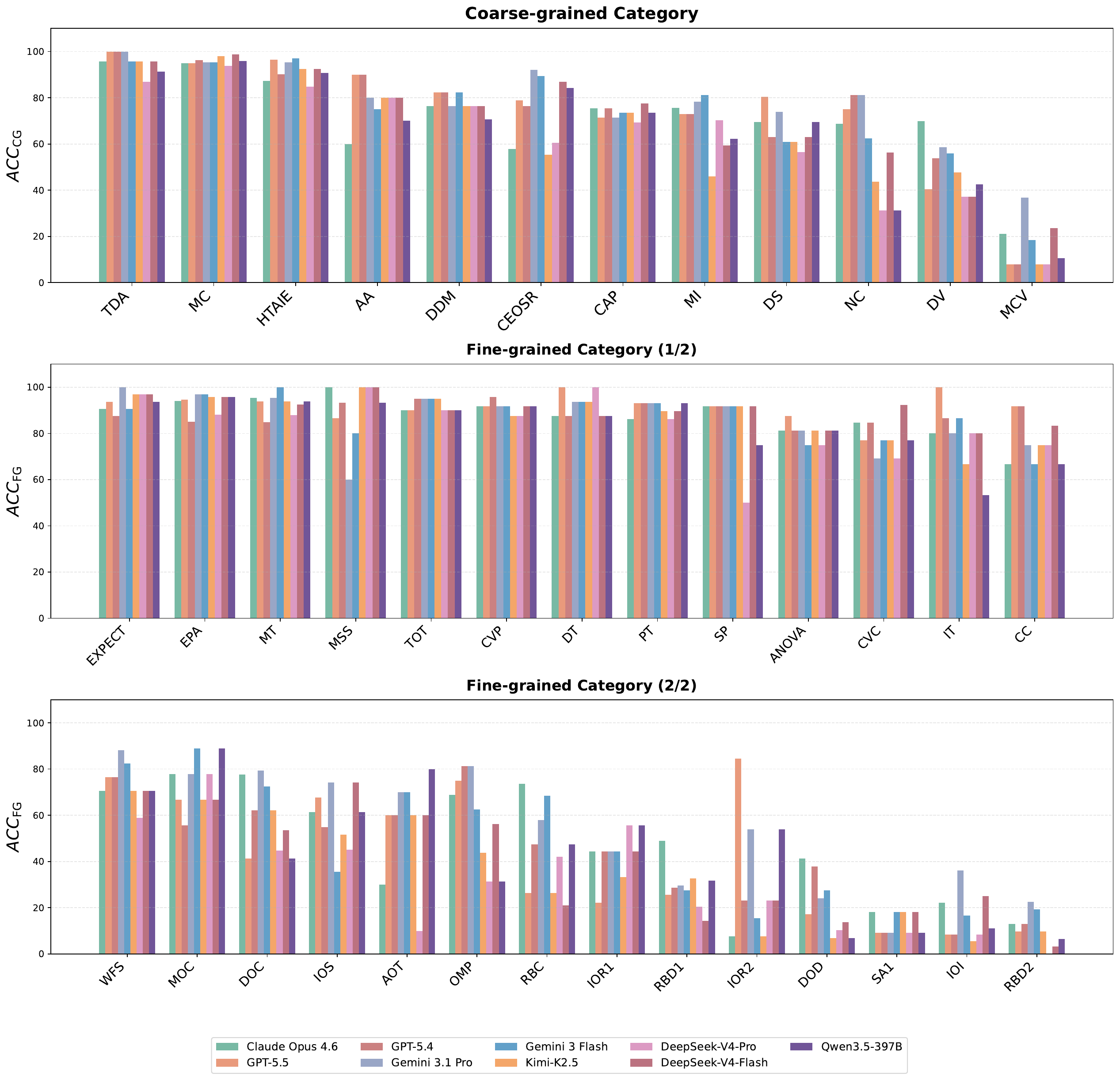}
\caption{Classification accuracy for all categories with more than 10 samples. Top: coarse-grained categories. Middle and bottom: fine-grained categories.}
\label{fig:acc_by_category_all}
\end{figure}

Figure~\ref{fig:acc_by_category_all} complements the category-level analysis with results for all categories containing more than 10 samples across ten models under the same evaluation setting. In line with the findings in Section~\ref{sec:fine_grained}, performance is higher when the analytical objective is explicit and lower when classification depends on subtle distinctions between related analytical intents.

\end{document}